\documentclass[conference]{IEEEtran}
\IEEEoverridecommandlockouts
\usepackage[compress]{cite}
\usepackage{amsmath,amssymb,amsfonts}
\usepackage{algorithm}
\usepackage[noend]{algpseudocode}
\usepackage{graphicx}
\usepackage{textcomp}
\usepackage{xcolor}
\usepackage[a4paper, total={184mm,239mm}]{geometry}
\usepackage{caption}
\usepackage{subcaption}
\usepackage{booktabs}
\usepackage{multirow}
\usepackage{url}
\usepackage{tikz}
\usepackage{hyperref}
\newcommand*\circled[1]{\tikz[baseline=(char.base)]{
            \node[shape=circle,draw,inner sep=0.4pt] (char) {#1};}}
\def\BibTeX{{\rm B\kern-.05em{\sc i\kern-.025em b}\kern-.08em
    T\kern-.1667em\lower.7ex\hbox{E}\kern-.125emX}}
\newcommand{\tabincell}[2]{\begin{tabular}{@{}#1@{}}#2\end{tabular}}

\begin{document}

\title{Subgraph Extraction-based Feedback-guided Iterative Scheduling for HLS
}

\author{
\IEEEauthorblockN{Hanchen Ye\IEEEauthorrefmark{1}\IEEEauthorrefmark{5}, David Z. Pan\IEEEauthorrefmark{2}\IEEEauthorrefmark{4}, Chris Leary\IEEEauthorrefmark{3}, Deming Chen\IEEEauthorrefmark{1}, Xiaoqing Xu\IEEEauthorrefmark{4}}
\IEEEauthorblockA{
\IEEEauthorrefmark{1}\textit{University of Illinois Urbana-Champaign}, \IEEEauthorrefmark{2}\textit{The University of Texas at Austin}, \IEEEauthorrefmark{3}\textit{Google}, \IEEEauthorrefmark{4}\textit{X, the moonshot factory} \\
hanchen8@illinois.edu, dpan@ece.utexas.edu, leary@google.com, dchen@illinois.edu, xiaoqingxu@x.team}
\thanks{\IEEEauthorrefmark{5}Work was done when interning at X, the moonshot factory.}
\vspace{-10pt}
}

\maketitle

\begin{abstract}
This paper proposes ISDC, a novel feedback-guided iterative system of difference constraints (SDC) scheduling algorithm for high-level synthesis (HLS). ISDC leverages subgraph extraction-based low-level feedback from downstream tools like logic synthesizers to iteratively refine HLS scheduling. Technical innovations include: (1)~An enhanced SDC formulation that effectively integrates low-level feedback into the linear-programming (LP) problem; (2)~A fanout and window-based subgraph extraction mechanism driving the feedback cycle; (3)~A no-human-in-loop ISDC flow compatible with a wide range of downstream tools and process design kits (PDKs). Evaluation shows that ISDC reduces register usage by 28.5\% against an industrial-strength open-source HLS tool.
\end{abstract}

% \begin{IEEEkeywords}
% component, formatting, style, styling, insert
% \end{IEEEkeywords}

\section{Introduction}
\label{sec:intro}

Scheduling is one of the most important problems in high-level synthesis (HLS) that partitions a computation graph into multiple clock cycles under the given timing and resource constraints.
% Historically, many HLS scheduling algorithms have been proposed, encompassing list scheduling~\cite{parker1986maha}, force-directed scheduling~\cite{paulin1989force}, and ILP-based scheduling~\cite{hwang1991formal}.
In 2006, Cong and Zhang~\cite{cong2006efficient} proposed a scheduling algorithm based on a system of difference constraints (SDC) formulation, converting the scheduling problem into a linear programming (LP) problem that can be solved optimally in polynomial time. SDC scheduling marked an important milestone for HLS and has been widely adopted in HLS tools, including AMD Vitis HLS~\cite{vitishls2022userguide}, LegUp~\cite{legup2021document}, and Google XLS~\cite{xls2023github}.

Over the years, both industrial and academic HLS tools~\cite{vitishls2022userguide, legup2021document, xls2023github} have been relying on high-level intermediate representation (IR), such as LLVM IR~\cite{lattner2004llvm}, for timing analysis, area/resource analysis, and scheduling. In this context, the IR operations, such as integer additions and multiplications, are viewed as the fundamental elements to schedule against. Their delays and resources are pre-characterized in isolation through downstream tools, such as logic synthesizer~\cite{wolf2016yosys, brayton2010abc}, for the target technology library.
% Following this, the scheduling algorithm employs these pre-characterization models to estimate timing and resource utilization, subsequently imposing the associated design constraints. 
While this can capture some low-level characteristics of individual operations, it does not model further optimizations in downstream tools, such as logic resubstitution and rewriting, leading to estimations that are substantially different from the actual quality of results (QoR)~\cite{dai2018fast}.

\begin{figure}[t]
    \centering
    \includegraphics[width=0.95\linewidth]{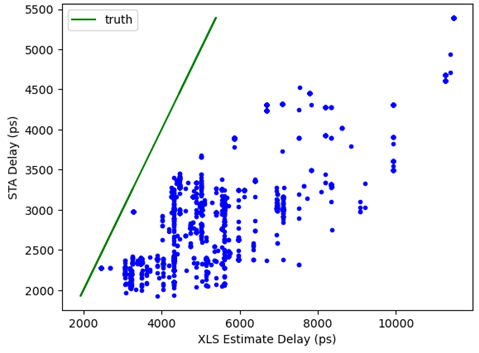}
    \vspace{-3pt}
    \caption{Post-synthesis STA vs. XLS-estimated critical path delay of 6912 different HLS design points.}
    \label{fig:xls_profiling}
    \vspace{-6pt}
\end{figure}

\begin{figure*}[t]
    \centering
    \includegraphics[width=0.95\textwidth]{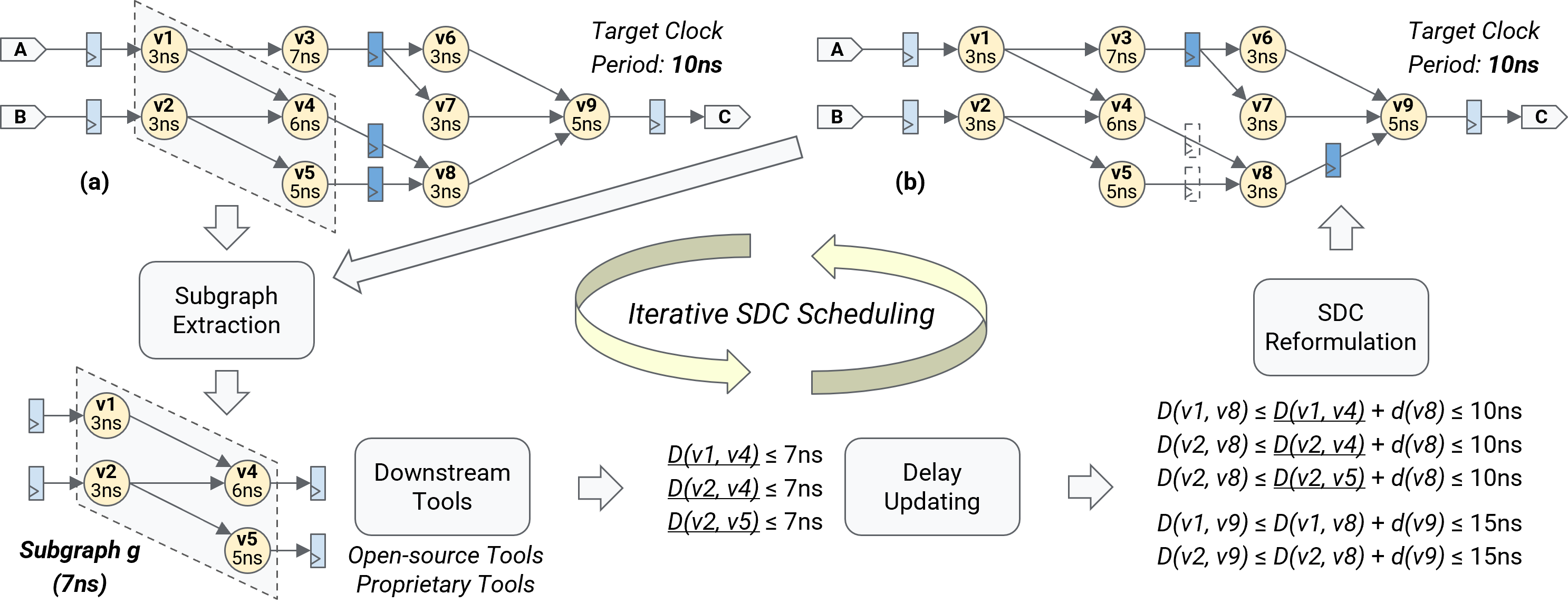}
    \vspace{-3pt}
    \caption{Overall flow of ISDC scheduling algorithm.}
    \label{fig:flow}
    \vspace{-6pt}
\end{figure*}

To study this phenomenon, we generated 6912 different design points of an HLS design with Google XLS~\cite{xls2023github} and profiled their post-synthesis static timing analysis (STA) and XLS-estimated critical path delays. We used Yosys~\cite{wolf2016yosys} and OpenSTA~\cite{opensta2023github} for logic synthesis and STA. We used SKY130~\cite{skywater2023github} as the target technology library. Fig.~\ref{fig:xls_profiling} shows the profiling results. We can observe the XLS-estimated delays (blue dots) exhibit significant deviation from the STA delays (the green line), which are treated as the ground truth for this experiment. These deviations create unused slack and present numerous opportunities to refine scheduling quality, such as reducing register usage. However, without access to low-level information, HLS tools cannot effectively capitalize on these opportunities.

Feedback-directed optimization (FDO) has been widely adopted in software compilation and shown significant benefits~\cite{chen2016autofdo}. However, this idea has not been thoroughly studied in the HLS domain.
Lucas~et~al.~\cite{lucas2009fastyield} presented a process variation and layout-aware HLS binding algorithm, which aimed to improve the performance yield of generated designs by enhancing the HLS binding process. Zheng~et~al.~\cite{zheng2014fast} introduced a placement and routing (PAR) directed HLS flow, which heavily depends on back annotations from a specific proprietary synthesis and PAR tool, limiting its adaptability to other scenarios. Tan~et~al.~\cite{tan2015mapping} presented a mixed integer linear programming (MILP) formulation for technology mapping-aware HLS scheduling. While it considers the mapping of original operations into look-up tables (LUTs), it cannot capture inter-operation optimizations present in logic synthesis and beyond. Rizzi~et~al.~\cite{rizzi2023iterative} introduced an iterative technology mapping-aware register placement algorithm. Nonetheless, its focus remains primarily on the LUT-mapping of FPGA targets and dynamically scheduled dataflow circuits.

In this paper, we introduce low-level feedback into the HLS scheduling problem and propose ISDC, a novel iterative SDC scheduling method. The main contributions are:
\begin{itemize}
    \item A scheduling algorithm that leverages feedback from downstream tools to refine the scheduling result iteratively and reduce register usage.
    \item An enhanced SDC formulation that effectively integrates low-level feedback into the LP problem.
    \item A fanout-driven and window-based subgraph extraction method that improves the quality and convergence speed of iterative scheduling.
    \item A no-human-in-loop ISDC workflow compatible with a wide range of downstream tools and PDK. ISDC is fully open-sourced at \url{https://github.com/google/xls}.
\end{itemize}

\section{Preliminaries}
\label{sec:prelims}

The HLS IR to be scheduled is typically represented as a directed graph $G$. For each operation node $v$ in graph $G$, SDC scheduling~\cite{cong2006efficient} defines a variable $s_v$ to represent the time step in which the operation is scheduled into. By ensuring constraints in integer-difference form, such as:
\begin{equation}
    s_u - s_v \le d_{u,v}
\end{equation}
where $d_{u,v}$ is an integer, a totally unimodular constraint matrix is derived, which is guaranteed to have integral solutions~\cite{zhang2013sdc}. A set of common HLS constraints can be expressed in the form of integer-difference constraints~\cite{zhang2013sdc}. Specifically, to meet the target clock frequency, a timing constraint is used to constrain the maximum combinational delay within a clock cycle. For the \emph{critical combinational path (CCP)} connecting $v_{i_1}$ and $v_{i_k}$ with the largest delay, we can calculate its delay $D(ccp(v_{i_1} , v_{i_k}))$ as $\sum^k_{s=1}d(v_{i_s})$, where $d(v)$ is the individual delay of $v$. For each operation pair $v_i$ and $v_j$ with $D(ccp(v_i, v_j)) > T_{clk}$, where $T_{clk}$ is the target clock period, we construct a constraint as:
\begin{equation}
\label{eq:sdc_timing_constraint}
    s_{v_i}-s_{v_j} \le -\left( \left\lceil \frac{D(ccp(v_i, v_j))}{T_{clk}}\right\rceil -1\right)
\end{equation}
Eq.~\ref{eq:sdc_timing_constraint} states that the combinational path with total delay exceeding the target clock period $T_{clk}$ must be partitioned into at least $\left\lceil D(ccp(v_i, v_j))/T_{clk}\right\rceil$ number of clock cycles.

\section{Iterative SDC Scheduling}

\subsection{Overall Flow}
\label{sec:overall_flow}

Fig.~\ref{fig:flow} shows the overall flow of the proposed ISDC scheduling algorithm. ISDC starts from an initial pipeline, as depicted in Fig.~\ref{fig:flow}(a), scheduled with the original SDC scheduling algorithm~\cite{cong2006efficient}. Note that each node in Fig.~\ref{fig:flow}(a) represents an operation of the HLS IR, such as integer additions and multiplications. On top of this initial schedule, a set of combinational subgraphs, such as subgraph $g$ at the lower-left of Fig.~\ref{fig:flow}, are extracted and passed to downstream tools for subgraph logic synthesis and beyond. Subsequently, the subgraph delays fed back from downstream tools are integrated into an enhanced SDC formulation to construct an updated LP problem. Upon solving this LP problem, a new pipeline schedule is generated as depicted in Fig~\ref{fig:flow}(b). This procedure is then iteratively applied to the new pipeline schedule until a stable scheduling result is achieved, exemplified by metrics such as register usage.

\subsubsection{Why low-level feedback helps}
As shown in Fig.~\ref{fig:flow}(a), the initial estimation of $D(ccp(v_2, v_8))$ is calculated as $d(v_2) + d(v_4) + d(v_8)$, which totals to 12ns. Given the target clock period of 10ns, $v_2$ and $v_8$ must be scheduled into separate clock cycles. However, suppose the delay of subgraph $g$ reported by downstream tools is 7ns, $D(ccp(v_2, v_8))$ can be recalculated as $D(g) + d(v_8)$, equaling to 10ns. As a result, $v_8$ can now be merged into the same clock cycle as $v_2$, leading to a decrease in register usage as depicted in Fig.~\ref{fig:flow}(b). This underscores the significance of low-level feedback in refining scheduling result. Such feedback empowers ISDC to identify better design points that might have been erroneously overlooked by the original SDC scheduling algorithm.

\subsubsection{Why an iterative approach helps}
Considering the real-world constraints of computational resources, it is infeasible to evaluate every subgraph in an HLS design for feedback, especially given the exponential increase in complexity as the HLS design grows. By using an iterative approach, ISDC can capitalize on knowledge from prior iterations, substantially reducing the search space of subgraph extraction by focusing on \emph{combinational} subgraphs from the previous schedule. This approach helps ISDC incrementally refine the scheduling result, maintaining manageable computational complexity throughout each iteration.

\subsection{Subgraph Extraction Strategy}

Despite using an iterative approach, the number of subgraph candidates remains vast, which can readily result in slow convergence. In this section, we introduce two orthogonal strategies to address this problem. Their ablation studies are presented in Section~\ref{sec:ablation_study}.

\begin{figure}[t]
    \centering
    \includegraphics[width=0.95\linewidth]{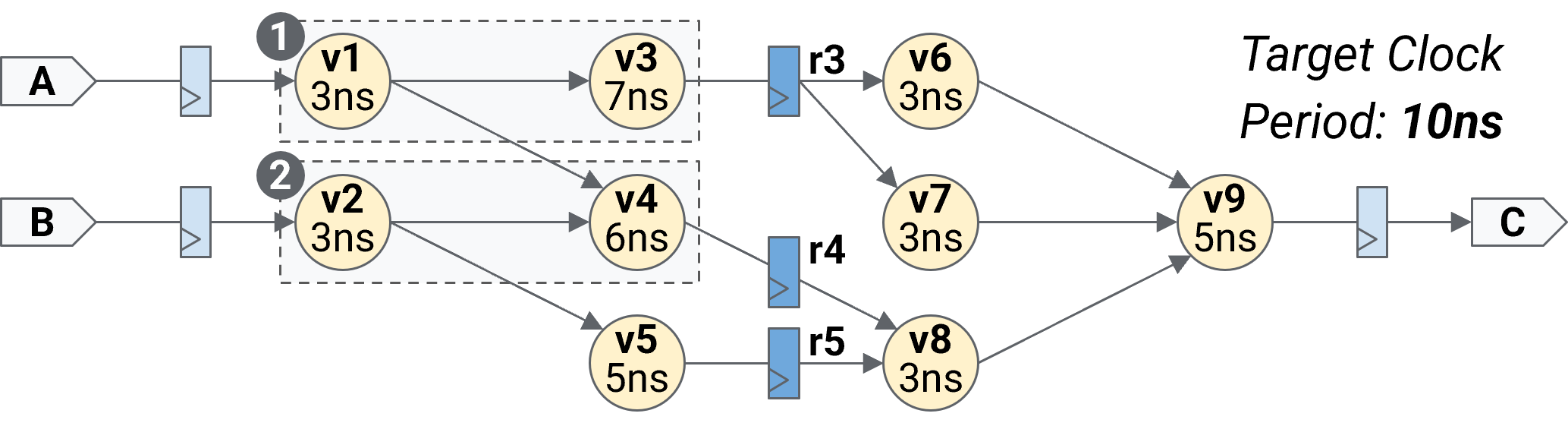}
    \vspace{-3pt}
    \caption{Delay-based vs. fanout-based subgraph extraction.}
    \label{fig:fanout_driven}
    \vspace{-6pt}
\end{figure}

\begin{figure}[t]
    \centering
    \includegraphics[width=0.9\linewidth]{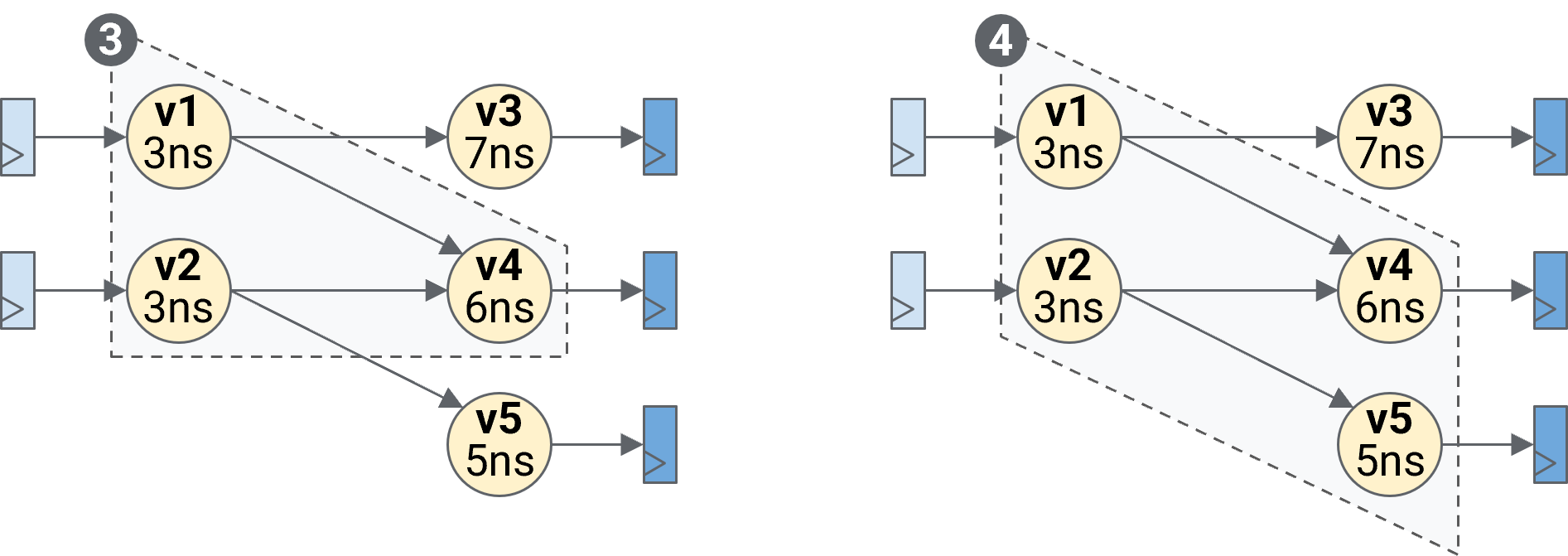}
    \vspace{-3pt}
    \caption{Cone-based vs. window-based subgraph extraction.}
    \label{fig:cone_window}
    \vspace{-6pt}
\end{figure}

\subsubsection{Fanout-driven strategy}
\label{sec:fanout_driven}
A direct and intuitive extraction strategy is delay-driven: focusing on the longest paths from the previous schedule because of their impact on the achievable clock frequency. Nonetheless, we argue that relying solely on delay is not the most effective strategy. As illustrated in Fig.~\ref{fig:fanout_driven}, path~\circled{1} is the longest combinational path with a delay of 10ns. But the register associated with path~\circled{1}, specifically $r_3$, is utilized by two consumer nodes, $v_6$ and $v_7$. Merging the two nodes into the first clock cycle would increase register usage, being not beneficial. In comparison, although path~\circled{2} has a shorter delay of 9ns, its associated register $r_4$ only has a single consumer, offering greater flexibility in its positioning.

Essentially, the more a register is being utilized, the more critical it becomes, reducing the benefit of repositioning it. Therefore, we introduce the following metric to drive the subgraph extraction process:
\begin{equation}
    S(v_i, v_j) = \sum^k_{s=1}\frac{bit\_count(r_s(v_j)) + \frac{D(ccp(v_i, v_j))}{T_{clk}}}{num\_users(r_s(v_j))+1}
\end{equation}
Assuming $v_j$ produces a total of $k$ results, $r_s(v_j)$ denotes the $s$-th result of $v_j$. The function $bit\_count$ quantifies the significance of $r_s(v_j)$, while $num\_users$ captures the degree to which $r_s(v_j)$ is utilized. $D(ccp(v_i, v_j))/T_{clk}$ serves as a tie-breaker, and is ensured to be less than $1.0$ in any valid schedule. Suppose $m$ subgraphs are extracted in each iteration, ISDC sorts all combinational paths from the previous schedule in descending order of $S(v_i, v_j)$ and extract the top $m$ paths. Given that $num\_users$ can be viewed as the HLS IR level fanout, we term this approach the \emph{fanout-driven} strategy.

\subsubsection{Window-based strategy}
The motivation of introducing feedback is to capture the low-level optimizations in downstream tools. To better capture inter-node optimizations, ISDC expands the paths identified in Section~\ref{sec:fanout_driven} to \emph{cone}s and \emph{window}s. Here, a cone is defined as a set of nodes at the HLS IR level with multiple input nodes (\emph{leaves}) and a single output node (\emph{root}). A cone must adhere to the following properties: (1)~Each path from any primary input (PI) of graph $G$ to $root$ passes though a leaf; (2)~For each leaf, there exists a path from a PI to $root$ that passes though that specific leaf and bypasses any other leaves. To expand a given path between nodes $v_i$ and $v_j$ into a combinational cone, ISDC uses a depth-first search (DFS) algorithm that recursively identifies the preceding nodes of $v_j$ until it encounters the boundary nodes of clock cycles or the PI of the entire graph $G$.

A window is derived by merging multiple cones that have different roots but share an identical or overlapping set of leaves. While a window still adheres to the properties above, it extends them to the case of multiple output nodes. Fig.~\ref{fig:cone_window} shows an example of expanding path~\circled{2} in Fig.~\ref{fig:fanout_driven} to a cone (subgraph~\circled{3}) and a window (subgraph~\circled{4}). Given that cone/window-based optimizations are prevalent in logic synthesis~\cite{brayton2010abc}, the cone/window subgraphs can capture the most relevant inter-operation optimizations, while also being sufficiently self-contained to mitigate the potential of over-optimization.

\setlength{\textfloatsep}{14pt}
\begin{algorithm}[t]
\small
\captionsetup{font=small}
\caption{Pseudo code of delay updating}
\label{alg:delay_updating}
\begin{algorithmic}[1]
\Require $G$, the graph to be scheduled
\Require $S[m]$, all evaluated subgraphs
\Require $D[n][n]$, the past critical path delay of all node pairs
\Ensure Updated $D[n][n]$, the critical path delay of all node pairs

\If{is\_first\_iteration()} \Comment{Initialize $D$}
\For{$u$ \textbf{in} get\_nodes($G$)}
\For{$v$ \textbf{in} get\_nodes($G$)}
\If{$u == v$}
\State $D[u][v] \gets d(v)$ \Comment{Get individual delay}
\ElsIf{is\_connected($u$, $v$)}
\State $D[u][v] \gets D(ccp(u, v))$ \Comment{Get critical path delay}
\Else
\State $D[u][v] \gets -1$ \Comment{Annotate as not connected}
\EndIf
\EndFor
\EndFor
\EndIf

\For{$g$ \textbf{in} $S$} \Comment{Traverse all evaluated subgraphs}
\For{$u$ \textbf{in} get\_nodes($g$)}
\For{$v$ \textbf{in} get\_nodes($g$)}
\If{$D[u][v] >$ get\_delay($g$) \textbf{and} $D[u][v] \ne -1$}
\State $D[u][v] \gets$ get\_delay($g$) \Comment{Update critical delay}
\EndIf
\EndFor
\EndFor
\EndFor
\end{algorithmic}
\end{algorithm}

\subsection{Delay Updating}

In the initial SDC scheduling phase, ISDC employs the method outlined in Section~\ref{sec:prelims} to calculate the critical path delay for every node pair and set timing constraints. To integrate the low-level feedback into the subsequent SDC formulations, ISDC maintains a matrix $D[n][n]$ that holds the estimated critical path delay of all node pairs, where $n$ denotes the total node count. In each iteration, ISDC updates $D[n][n]$ with Alg.~\ref{alg:delay_updating} once the subgraph delays are fed back from downstream tools. Lines 1 to 9 initialize $D[n][n]$ with the naive delay estimations. Subsequently, lines 10 to 14 traverse all evaluated subgraphs. For each subgraph $g$, the delay of all node pairs covered by $g$ is updated with $D(g)$, but only if $D(g)$ is shorter than the original delay estimation. Consequently, ISDC maximally leverages the information obtained from each subgraph evaluation, thereby accelerating the iterative convergence.

\begin{algorithm}[t]
\small
\captionsetup{font=small}
\caption{Pseudo code of SDC reformulation}
\label{alg:sdc_reformulation}
\begin{algorithmic}[1]
\Require $G$, the graph to be scheduled
\Require $T_{clk}$, target clock period
\Require Updated $D[n][n]$, the critical path delay of all node pairs
\Ensure $M$, the reformulated SDC model

\State $V \gets$ get\_nodes($G$)
\For{$v$ \textbf{in} topo\_sort($V$)}
\State $D_v[n] \gets$ new($[-1] \times n$)
\For{$p$ \textbf{in} get\_operands($v$)} \Comment{Traverse all operands of $v$}
\For{$u$ \textbf{in} $V$}
\If{$D[u][p] \ne -1$}
\If{$D_v[u] < D[u][p] + D[v][v]$}
\State $D_v[u] \gets D[u][p] + D[v][v]$
\EndIf
\EndIf
\EndFor
\EndFor
\For{$u$ \textbf{in} $V$}
\If{$D_v[n] \ne -1$}
\If{$D[u][v] > D_v[n]$ \textbf{or} $D[u][v] == -1$}
\State $D[u][v] \gets D_v[n]$ \Comment{Update critical path delay}
\EndIf
\EndIf
\EndFor
\EndFor

\For{$u$ \textbf{in} reverse\_topo\_sort($V$)}
\State $D_u[n] \gets$ new($[-1] \times n$)
\For{$c$ \textbf{in} get\_users($u$)} \Comment{Traverse all users of $u$}
\EndFor
\State ... ... \Comment{Reversed delay propagation}
\EndFor

\State $M \gets$ initialize\_sdc($V$)
\For{$u$ \textbf{in} $V$}
\For{$v$ \textbf{in} $V$}
\If{$D[u][v] > T_{clk}$}
\State add\_timing\_constraint($M$) \Comment{Set Eq.~\ref{eq:sdc_timing_constraint} to $M$}
\EndIf
\EndFor
\EndFor
\State add\_other\_constraint($M$)
\end{algorithmic}
\end{algorithm}

\subsection{SDC Reformulation}
Upon the updated delay matrix $D[n][n]$, all the timing constraints discussed in Section~\ref{sec:prelims} are reformulated to construct an updated LP problem. Essentially, this reformulation can be viewed as an all-pairs shortest path problem, optimally solved by Floyd-Warshall algorithm with a complexity of $O(n^3)$. To mitigate this cubic complexity, we propose an $O(n^2)$ algorithm as Alg.~\ref{alg:sdc_reformulation}, which provides a sufficiently accurate delay estimation for our purposes. The estimation accuracy study is presented in Section~\ref{sec:benchmarking}. Lines 2 to 12 traverse all nodes of graph $G$ in a topological order, ensuring that a node is processed only after all its operand nodes. For a specific node $v$, lines 4 to 8 calculate the delay from all nodes to $v$ by adding $v$'s individual delay to the delay from all nodes to $v$'s operand nodes. Lines 7 to 8 ensure only the critical path delay is recorded. Subsequently, lines 9 to 12 update $D[n][n]$ only if the newly calculated delay is shorter.

After this topological order traversal, lines 13 to 16 of Alg.~\ref{alg:sdc_reformulation} reprocess all nodes, but in a reversed topological order. This step aims to identify the complementary paths that cannot be identified by the initial topological order traversal. Finally, lines 18 to 21 set the timing constraints for the LP problem based on the recalculated $D[n][n]$. Intuitively, by reformulating the SDC problem, ISDC prunes the over-conservative timing constraints that were erroneously set in the previous SDC scheduling. This enlarges the updated LP problem's search space, naturally leading to a refined scheduling result.
% Given that typical critical path algorithms also have a worst-case complexity of $O(n^2)$, our proposed SDC reformulation algorithm doesn't notably increase the complexity of the original SDC scheduling.

\section{Evaluation}
\label{sec:evaluation}

We implemented the proposed ISDC algorithm on top of an industrial-strength open-source HLS tool, Google XLS~\cite{xls2023github}, which uses SDC scheduling~\cite{cong2006efficient} as the default scheduling algorithm. We used Yosys~\cite{wolf2016yosys} and OpenSTA~\cite{opensta2023github} for logic synthesis and STA. We used open-source SKY130~\cite{skywater2023github} as the target technology library.

\subsection{Ablation Study}
\label{sec:ablation_study}

We performed a set of ablation studies on an XLS-based HLS design to demonstrate the efficacy of the proposed fanout-driven and window-based subgraph extraction strategy.

\begin{figure*}[t]
    \centering
    \begin{subfigure}[b]{0.33\textwidth}
        \centering
        \includegraphics[width=\textwidth]{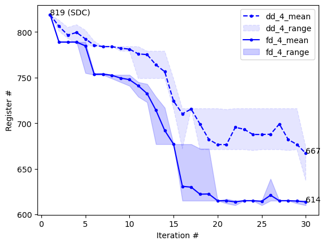}
        \vspace{-15pt}
        \caption{4 subgraphs per iteration.}
    \end{subfigure}%
    \hfill % This command inserts a space between the subfigures. Remove it if you don't want any space.
    \begin{subfigure}[b]{0.33\textwidth}
        \centering
        \includegraphics[width=\textwidth]{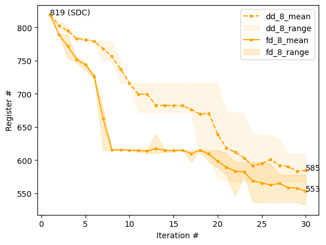}
        \vspace{-15pt}
        \caption{8 subgraphs per iteration.}
    \end{subfigure}%
    \hfill % This command inserts a space between the subfigures. Remove it if you don't want any space.
    \begin{subfigure}[b]{0.33\textwidth}
        \centering
        \includegraphics[width=\textwidth]{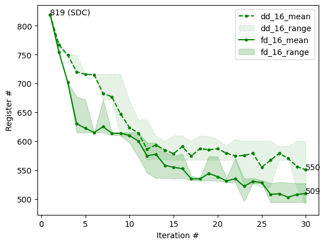}
        \vspace{-15pt}
        \caption{16 subgraphs per iteration.}
    \end{subfigure}
    \vspace{-2pt}
    \caption{Ablation study of delay-driven and fanout-driven subgraph extraction. Path-based strategy is used.}
    \label{fig:fanout_driven_results}
    \vspace{-3pt}
\end{figure*}

\subsubsection{Fanout-driven strategy}
Fig.~\ref{fig:fanout_driven_results} shows the comparisons between the delay-driven (dd) and fanout-driven (fd) strategies. We performed 30 iterations of scheduling under 400MHz clock frequency, where 4, 8, or 16 subgraphs were extracted per iteration. The results indicate that the fanout-driven strategy converges notably faster than its delay-driven counterpart, particularly in the initial iterations. Furthermore, it consistently achieves lower register usage across all three cases.
% The curve ranges in Fig.~\ref{fig:fanout_driven_results} also suggest that the fanout-driven strategy offers more stable performance compared to the delay-driven approach.
% However, due to the limitation of the path-based strategy, none of the experiments in this study reached a scheduling result with optimal register usage.

\begin{figure*}[t]
    \centering
    \begin{subfigure}[b]{0.33\textwidth}
        \centering
        \includegraphics[width=\textwidth]{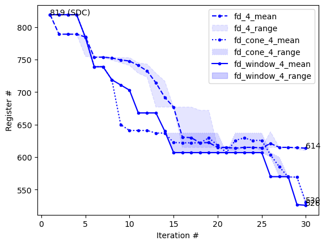}
        \vspace{-15pt}
        \caption{4 subgraphs per iteration.}
    \end{subfigure}%
    \hfill % This command inserts a space between the subfigures. Remove it if you don't want any space.
    \begin{subfigure}[b]{0.33\textwidth}
        \centering
        \includegraphics[width=\textwidth]{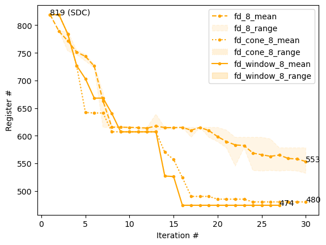}
        \vspace{-15pt}
        \caption{8 subgraphs per iteration.}
        \label{subfig:cone_window_results_8}
    \end{subfigure}%
    \hfill % This command inserts a space between the subfigures. Remove it if you don't want any space.
    \begin{subfigure}[b]{0.33\textwidth}
        \centering
        \includegraphics[width=\textwidth]{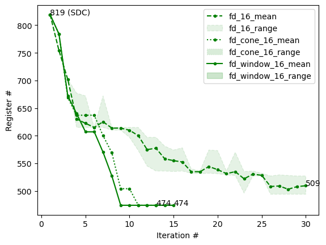}
        \vspace{-15pt}
        \caption{16 subgraphs per iteration.}
    \end{subfigure}
    \vspace{-2pt}
    \caption{Ablation study of path, cone, and window-based subgraph extraction. Fanout-driven strategy is used because it has produced better results than the delay-driven strategy as shown in Fig.~\ref{fig:fanout_driven_results}.}
    \label{fig:cone_window_results}
    \vspace{-6pt}
\end{figure*}

\subsubsection{Window-based strategy}
Fig.~\ref{fig:cone_window_results} shows the comparisons among the path, cone, and window-based strategies. Notably, the cone/window-based strategies demonstrate faster convergence than the path-based approach, achieving a reduced register usage. Path-based strategy is often trapped in local minima. In contrast, the cone/window-based strategy can overcome those points, achieving further improvements in subsequent iterations. While the cone and window-based strategies exhibit similar performance, the results suggest a slight edge for the window-based approach. Meanwhile, as expected, ISDC converges faster with the extraction and evaluation of more subgraphs per iteration.

% \begin{figure}[t]
%     \centering
%     \includegraphics[width=\linewidth]{figures/frequency.png}
%     \caption{Study of various frequencies.}
%     \label{fig:frequency}
% \end{figure}

% \subsubsection{Clock frequency}
% Fig.~\ref{fig:frequency}

\begin{table*}[t]
    \small
    \centering
    \caption{Benchmarking results on 17 XLS-based HLS designs.}
    \label{tab:results}
    \setlength\tabcolsep{4pt}
    \begin{tabular}{ccccccccccc}
    \toprule
    \multirow{2}{*}[-8pt]{\textbf{Benchmark}} &
    \multirow{2}{*}[-2pt]{\tabincell{c}{\textbf{Clock} \\ \textbf{Period} \\ \textbf{(ps)}}} &
    \multicolumn{4}{c}{\textbf{XLS~\cite{xls2023github} (SDC Scheduling)}} &
    \multicolumn{5}{c}{\textbf{Ours (Iterative SDC Scheduling)}} \\
    \cmidrule(lr){3-6} \cmidrule(lr){7-11}
    & & \tabincell{c}{\textbf{Slack} \\ \textbf{(ps)}} &
    \tabincell{c}{\textbf{Stage} \\ \textbf{Num.}} &
    \tabincell{c}{\textbf{Register} \\ \textbf{Num.}} &
    \tabincell{c}{\textbf{Schedule} \\ \textbf{Time (s)}} &
    \tabincell{c}{\textbf{Slack} \\ \textbf{(ps)}} &
    \tabincell{c}{\textbf{Stage} \\ \textbf{Num.}} &
    \tabincell{c}{\textbf{Register} \\ \textbf{Num.}} &
    \tabincell{c}{\textbf{Schedule} \\ \textbf{Time (s)}} &
    \tabincell{c}{\textbf{Iteration} \\ \textbf{Num.}} \\
    \midrule
    ML-core datapath1 & 2500 & 1161.65 & 2 & 99 & 0.14 & 729.72 & 1 & 50 & 6.73 & 3 \\
    ML-core datapath0 opcode4 & 5000 & 943.93 & 2 & 109 & 0.11 & 943.93 & 2 & 109 & 0.10 & 1 \\
    rrot & 2500 & 866.23 & 2 & 192 & 0.08 & 499.33 & 1 & 96 & 2.98 & 2 \\
    ML-core datapath0 opcode3 & 5000 & 1440.65 & 3 & 138 & 0.13 & 772.87 & 2 & 101 & 23.90 & 6 \\
    binary\_divide & 2500 & 518.66 & 3 & 71 & 0.12 & 436.18 & 3 & 70 & 7.56 & 4 \\
    hsv2rgb & 5000 & 1450.73 & 3 & 134 & 0.11 & 1149.73 & 2 & 102 & 10.64 & 3 \\
    ML-core datapath0 opcode0 & 5000 & 1140.9 & 3 & 162 & 0.12 & 1162.66 & 2 & 108 & 19.26 & 4 \\
    crc32 & 2500 & 1744.35 & 3 & 75 & 0.11 & 1686.49 & 1 & 38 & 4.76 & 3 \\
    ML-core datapath0 opcode1 & 5000 & 1235.58 & 5 & 298 & 0.15 & 1519.2 & 4 & 234 & 21.28 & 4 \\
    ML-core datapath0 opcode2 & 5000 & 1331.25 & 6 & 480 & 0.44 & 1030.73 & 3 & 209 & 94.30 & 14 \\
    ML-core datapath0 (all opcodes) & 5000 & 1834.68 & 8 & 1214 & 1.62 & 951.24 & 5 & 729 & 101.61 & 13 \\
    ML-core datapath2 & 2500 & 220.14 & 10 & 819 & 0.43 & 36.71 & 6 & 474 & 27.62 & 9 \\
    float32\_fast\_rsqrt & 5000 & 1202.02 & 10 & 1055 & 1.79 & 144.91 & 8 & 797 & 118.47 & 14 \\
    video-core datapath & 2500 & 26.86 & 12 & 1756 & 24.28 & 166.31 & 12 & 1732 & 316.62 & 11 \\
    internal datapath & 2500 & 371.22 & 26 & 3095 & 13.73 & 60.42 & 25 & 2976 & 167.04 & 10 \\
    sha256 & 2500 & 232.66 & 112 & 85545 & 284.47 & 74.11 & 97 & 73990 & 3280.88 & 11 \\
    fpexp\_32 & 5000 & 442.75 & 121 & 30569 & 240.90 & 236.97 & 114 & 29242 & 3441.08 & 13 \\
    \midrule
    \textbf{Geo. Mean} & & 686.74 & 6.93 & 569.86 & 0.84 & 418.16 & 4.85 & 407.19 & 34.46 & \\
    \textbf{Ratio} & & 100.0\% & 100.0\% & 100.0\% & 100.0\% & \textbf{60.9\%} & \textbf{70.0\%} & \textbf{71.5\%} & \textbf{4080.5\%} & \\
    \bottomrule
    \end{tabular}
    \vspace{-5pt}
\end{table*}

\begin{figure}[t]
    \centering
    \includegraphics[width=0.95\linewidth]{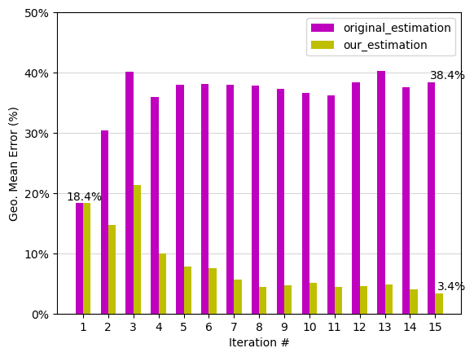}
    \vspace{-3pt}
    \caption{Delay estimation accuracy comparison.}
    \label{fig:accuracy}
    \vspace{-6pt}
\end{figure}

\subsection{Benchmarking Results}
\label{sec:benchmarking}

We performed benchmarking on 17 XLS-based HLS designs to evaluate ISDC. The benchmarks encompass common algorithms like \emph{crc32}, as well as datapaths from industrial SoCs, including an machine learning processor (\emph{ML-core}) and a video processor (\emph{video-core}). In the evaluation, we used the fanout-driven and window-based strategy, evaluating 16 subgraphs per iteration in parallel. A total of 15 iterations were performed on each benchmark. Tab.~\ref{tab:results} shows the evaluation results, including metrics such as the target clock period, post-synthesis slack, number of pipeline stages, number of registers, and scheduling runtime. By default, we set the target clock period to 2500ps to constrain the scheduling. If an operation in a benchmark exhibited an individual delay exceeding 2500ps, we adjusted the target clock period to 5000ps. On average, ISDC achieves a 28.5\% lower register usage compared to the original SDC scheduling. This comes at the cost of an average 40.8$\times$ increase in scheduling runtime. For instance, for the largest benchmark, \emph{sha256}, ISDC spends around 54.7 minutes to converge, which is 11.5$\times$ longer than the original SDC's 4.7 minutes. Among all benchmarks, ISDC utilized 39.1\% additional slack in average to make room for the reduction in register usage. However, there are several counter examples, such as \emph{ML-core datapath0 opcode0}, which exhibits a slight increase in slack but still achieves a register usage reduction.

To evaluate ISDC's delay estimation accuracy, we analyzed its estimation across the 17 benchmarks and compared with the original SDC. Fig.~\ref{fig:accuracy} shows the comparison results. In the first iteration, without low-level feedback, ISDC exhibits the same estimation error as the original SDC. However, as the iterations progress, ISDC gradually reduces its estimation error, ultimately reaching an error of 3.4\%. In contrast, the original SDC's estimation error increases. We attribute this to the fact that as the scheduling results are refined, more low-level optimizations are overlooked by the original SDC.

\section{Discussion}

\subsubsection{Process node}
Though we used real-world benchmarks for evaluation, we evaluated them down-clocked and on an older open-source industry process node (SKY130) to pioneer the methodology. We expect that the improvements should apply as effectively to more advanced process nodes and proprietary tools that offer similar STA report facilities.

\subsubsection{Retiming}
Retiming~\cite{leiserson1991retiming} is a method that repositions registers in gate-level sequential circuits to optimize performance or reduce resource usage without altering the overall functionality. On the other hand, HLS scheduling operates at higher-level IRs composed with algebraic operations and explicit control flows. This provides HLS scheduling with greater flexibility and larger design space to find more optimized designs. Furthermore, HLS scheduling preserves the algebraic attributes in the generated circuits, paving the way for robust verification processes, such as logic equivalence checking. This mitigates the limitations inherent in the retiming technique.

\begin{figure}[t]
    \centering
    \includegraphics[width=0.95\linewidth]{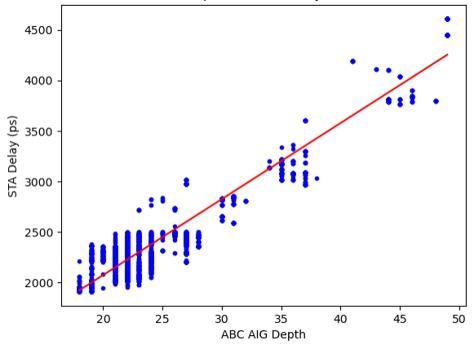}
    \vspace{-3pt}
    \caption{Post-synthesis STA vs. ABC~\cite{brayton2010abc} AIG depth of 6912 different HLS design points.}
    \label{fig:aig}
    \vspace{-6pt}
\end{figure}

\subsubsection{Runtime}
A common concern of feedback-guided approaches is runtime. While the results in Tab.~\ref{tab:results} demonstrate that ISDC converges at a practical pace, we have also explored a more aggressive strategy using the and-inverter-graph (AIG) to guide the scheduling. AIG is a widely adopted representation for logic optimizations in tools like ABC~\cite{brayton2010abc}. As shown in Fig.~\ref{fig:aig}, there is a compelling linear correlation between post-synthesis STA delay and AIG depth within ABC. This suggests a future research direction of bypassing the time-consuming technology mapping and post-synthesis STA, and directly using AIG depth as feedback.

\subsubsection{Simultaneous HLS and logic optimization}
In ISDC, we consciously bypass the back annotation technique used in~\cite{zheng2014fast, rizzi2023iterative} due to its backend-specific nature and lack of generalizability. However, to squeeze out the extra bit of performance from digital circuits in the post-Dennard-scaling era, it is possible to blur the lines between HLS and downstream processes, such as logic synthesis. Future endeavors might see a co-optimization of the two design spaces, such as simultaneous HLS scheduling and logic optimization.

\section{Conclusion}
In this paper, we proposed ISDC, a feedback-guided iterative scheduling algorithm for HLS. Building upon the traditional SDC approach, ISDC integrates iterative refinements and downstream tool feedback, showing a notable reduction in register usage.
% The addition of fanout-driven and window-based strategies enhances its adaptability.
ISDC offers insights into feedback-guided optimization for HLS and highlights avenues for future exploration.

% \section*{Acknowledgment}
% acknowledgement

\bibliographystyle{IEEEtran}
\bibliography{references}

\end{document}